\documentclass[11pt]{article}

\usepackage[preprint]{acl}
\usepackage{times}
\usepackage{latexsym}
\usepackage{enumitem}
\usepackage{amsmath}
\usepackage{amssymb}
\usepackage{longtable}
\usepackage{booktabs}
\usepackage{longtable}
\usepackage{booktabs}
\usepackage{xcolor}
\usepackage{array}

\usepackage[T1]{fontenc}
\usepackage[utf8]{inputenc}

\usepackage{microtype}

\usepackage{inconsolata}

\usepackage{graphicx}

\title{CLARA: Can AI Assess Developmental Appropriateness in Children’s Stories?}

\author{
\textbf{Sijing Yin\textsuperscript{1}},
\textbf{Zirui Wang\textsuperscript{1}},
\textbf{Qian Liu\textsuperscript{1}},
\textbf{Jiamou Liu\textsuperscript{1}\thanks{Corresponding author}}
\\[0.3em]
\textsuperscript{1}University of Auckland
\\[0.3em]
\small{
\texttt{
syin565@aucklanduni.ac.nz,
jiamou.liu@auckland.ac.nz
}
}
}

\begin{document}
\maketitle
\begin{abstract}

Assessing the developmental suitability of children’s narratives is important for educational recommendation and developmental literacy research, yet such assessment typically relies on subjective and difficult-to-scale human judgment. This raises an important question: \emph{Can AI systems approximate human developmental judgments of children’s stories?} To study this problem, we introduce CLARA, a cognitively grounded framework for developmental narrative understanding through structured annotation across cognitive (COG), language (LAN), and social-emotional (SEL) dimensions, together with a bilingual benchmark resource containing 1107 Chinese--English children’s stories with normalized silver developmental references and structured developmental annotations. We evaluate CLARA through benchmark comparison, component analysis, translated bilingual consistency analysis, and blinded human evaluation with educators. Experimental results show that structured developmental annotation achieves substantially stronger alignment with developmental references and human judgments than readability-based methods and direct prompting baselines. Overall, our findings suggest that AI systems can approximate certain aspects of human developmental judgment when guided by structured developmental annotation, while also highlighting the importance of interpretability and human oversight in educational NLP.

\end{abstract}

\section{Introduction}
\label{sec:intro}

Assessing the developmental suitability of children’s narratives is important for educational recommendation, curriculum design, and developmental literacy research. In practice, however, such assessment typically relies on human judgment from educators, parents, or publishers, which is often subjective, culturally dependent, and difficult to scale~\citep{gilardi2023chatgpt}. Existing computational approaches mainly estimate reading difficulty through readability formulas or direct age prediction, focusing on lexical and syntactic complexity rather than the developmental abilities required for narrative understanding~\citep{flesch1948new, kincaid1975derivation, dale1948formula}. Yet understanding children’s stories often requires substantially more than surface language processing, including causal reasoning, temporal understanding, emotional interpretation, social reasoning, and perspective taking~\citep{piaget1952origins, vygotsky1978mind, astington1988developing, wellman2004scaling}. Consequently, two stories with similar readability may still require very different levels of developmental understanding.

This raises an important question: \emph{Can AI systems approximate human developmental judgments of children’s stories?} More specifically, can language models identify the cognitive, linguistic, and social-emotional abilities required for understanding narratives in a way that aligns with human educational reasoning? Addressing this problem requires not only predictive models, but also structured developmental representations, interpretable annotation spaces, and cognitively grounded narrative analysis frameworks capable of modeling developmental narrative understanding beyond readability alone~\citep{rashkin2018naive, mostafazadeh2020glucose, vandijk2023theory}.
% structured developmental representations, interpretable annotation spaces, and evaluation protocols capable of measuring developmental narrative understanding beyond readability alone~\citep{rashkin2018naive, mostafazadeh2020glucose}.

To study this problem, we introduce CLARA (\textbf{C}ognitive \textbf{L}abeling and \textbf{A}ge \textbf{R}easoning for N\textbf{A}rratives), a cognitively grounded framework for developmental narrative understanding through structured annotation. Rather than directly predicting developmental stages from stories, CLARA decomposes developmental understanding into interpretable developmental representations across three dimensions: cognitive skills (COG), language abilities (LAN), and social-emotional abilities (SEL)~\citep{wei2022chain, wang2022self, zhou2023least}. The framework consists of four components: (1) a developmental annotation taxonomy, (2) a prompt-based annotation protocol, (3) a lightweight developmental aggregation mechanism, and (4) a bilingual benchmark resource for evaluation. Importantly, the current implementation uses GPT-4o as one possible annotation engine rather than a required component of the framework itself.

% To support systematic evaluation, we construct a bilingual benchmark containing 1107 Chinese--English children’s stories with normalized silver developmental references and structured developmental annotations. Existing narrative understanding resources mainly focus on story comprehension or commonsense reasoning rather than developmental cognitive representations~\citep{xu2022fairytaleqa, tikhonov2021storydb, mostafazadeh-etal-2016-corpus}. Experimental results show that CLARA substantially outperforms readability-based and direct prompting baselines, improving overlap accuracy from 0.603 (FKGL) and 0.730 (GPT-4o Direct) to 0.904 while reducing mean developmental distance to 0.096. Human evaluation further shows 0.81 agreement with educator developmental rankings and an interpretability score of 4.42/5. These findings suggest that developmental narrative understanding cannot be captured through surface linguistic complexity alone, but instead requires explicit modeling of cognitive, linguistic, and social-emotional developmental signals~\citep{rashkin2018naive, mostafazadeh2020glucose}.

To support systematic evaluation, we construct a bilingual benchmark containing 1107 Chinese--English children’s stories with normalized silver developmental references and structured developmental annotations. Here, ``bilingual'' refers to aligned Chinese--English versions of the same stories, where the English narratives are translated from the Chinese sources. Accordingly, the bilingual analysis evaluates consistency under translation rather than cross-lingual or cross-cultural generalization. Existing narrative understanding resources mainly focus on story comprehension or commonsense reasoning rather than developmental cognitive representations~\citep{xu2022fairytaleqa, tikhonov2021storydb, mostafazadeh-etal-2016-corpus}. Experimental results show that CLARA substantially outperforms readability-based and direct prompting baselines, improving overlap accuracy from 0.603 (FKGL) and 0.730 (GPT-4o Direct) to 0.904 while reducing mean developmental distance to 0.096. Human evaluation further shows 0.81 agreement with educator developmental rankings and an interpretability score of 4.42/5. These findings suggest that developmental narrative understanding cannot be captured through surface linguistic complexity alone, but instead requires explicit modeling of cognitive, linguistic, and social-emotional developmental signals~\citep{rashkin2018naive, mostafazadeh2020glucose}.

Overall, our findings suggest that AI systems can approximate certain aspects of human developmental judgment when guided by structured developmental annotation, while also highlighting the importance of interpretability and human oversight in educational NLP~\citep{gilardi2023chatgpt}.

\smallskip

\noindent\textbf{Contributions.}
\begin{itemize}[leftmargin=*]

\item We formulate developmental narrative understanding as a structured and interpretable NLP problem centered on approximating human developmental judgments of children’s stories.

\item We introduce CLARA, a cognitively grounded framework that models developmental understanding through structured annotations across cognitive, language, and social-emotional dimensions.

\item We construct a bilingual benchmark resource containing 1107 Chinese--English children’s stories with normalized silver developmental references and structured developmental annotations.

\item We conduct benchmark comparison, component analysis, translated bilingual consistency analysis, and blinded human evaluation, showing that structured developmental annotation aligns more closely with developmental references and human judgments than readability-based and direct prompting approaches.

\end{itemize}

\begin{figure*}[t]
\centering
\includegraphics[width=0.95\textwidth]{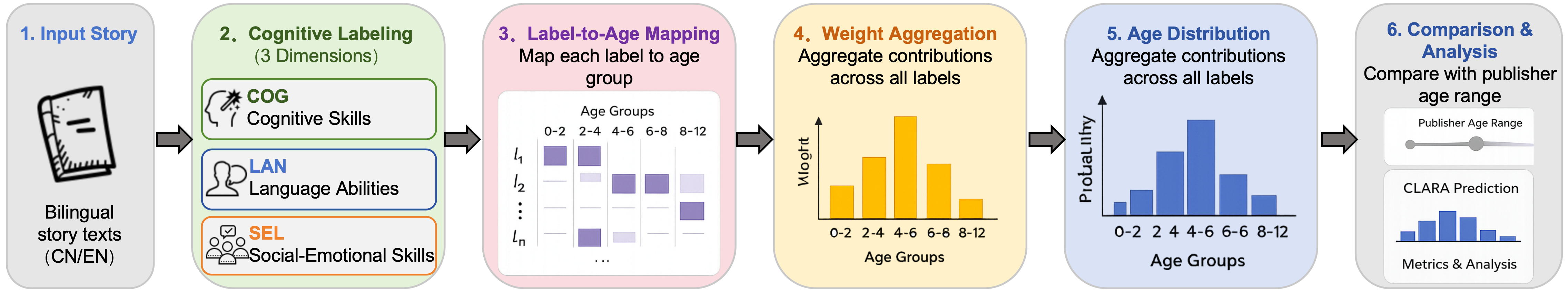}
\caption{Overview of the CLARA framework. Given bilingual story narratives, CLARA generates structured developmental annotations across COG, LAN, and SEL dimensions, aggregates developmental evidence into developmental-stage estimates, and compares the resulting representations with human and publisher developmental judgments.}
\label{fig:overview}
\end{figure*}

\section{Related Work}
\label{sec:related_work}

\subsection{Readability and Developmental Assessment}

Traditional approaches to reading difficulty and age estimation mainly rely on readability formulas~\citep{flesch1948new, kincaid1975derivation, dale1948formula}. 
Recent work further applies machine learning and neural language models to text readability and educational assessment~\citep{martinc2021readability, gilardi2023chatgpt}. 
However, these approaches mainly model surface linguistic complexity and provide limited developmental interpretability.

\subsection{Developmental Narrative Understanding}

Developmental psychology studies how children’s cognitive, linguistic, and social-emotional abilities evolve over time~\citep{piaget1952origins, vygotsky1978mind, casel2012framework}. Prior work on children’s narratives further highlights the importance of causal reasoning, perspective taking, emotional interpretation, narrative coherence, and discourse organization in developmental language understanding~\citep{astington1988developing, wellman2004scaling, reese2011coherence, gagarina2019main, gagarina2022referentiality}. Research on bilingual and developmental narratives additionally studies narrative macrostructure, referential coherence, and discourse-level developmental progression across children with different linguistic backgrounds~\citep{yang2022bilingual, kupersmitt2019causal}. Recent NLP-based studies further explore theory-of-mind reasoning and developmental signals in children’s narratives~\citep{vandijk2023theory}. However, these developmental frameworks are rarely operationalized into scalable computational annotation systems for NLP.

\subsection{Structured Story Understanding}

Narrative understanding has been widely studied in NLP, including story comprehension, commonsense reasoning, character-centered inference, and structured narrative explanation~\citep{mostafazadeh-etal-2016-corpus, rashkin2018naive, sap2019atomic, mostafazadeh2020glucose}. Several works further introduce structured cognitive annotations for stories, such as Story Commonsense~\citep{rashkin2018naive} and GLUCOSE~\citep{mostafazadeh2020glucose}. Educational narrative resources such as FairytaleQA~\citep{xu2022fairytaleqa} and StoryDB~\citep{tikhonov2021storydb} additionally connect narrative understanding with educational settings. Unlike prior work, CLARA studies whether AI systems can approximate human developmental judgments of children's narratives through structured cognitive, linguistic, and social-emotional annotation.

\section{Developmental Annotation Framework}
\label{sec:framework}

CLARA studies whether AI systems can approximate human developmental judgments of children's narratives through structured developmental annotation. 
Rather than directly predicting developmental stages, CLARA represents narratives through COG, LAN, and SEL developmental signals.

% CLARA consists of four components: 
% (1) a developmental taxonomy, 
% (2) a prompt-based annotation protocol, 
% (3) a lightweight aggregation mechanism, and 
% (4) a bilingual benchmark resource. 
% GPT-4o is used as one possible annotation engine rather than a required component of the framework. 
% The framework is model-agnostic and can be instantiated with different annotation models under the same developmental taxonomy and prompting protocol.

CLARA consists of four components: 
(1) a developmental taxonomy, 
(2) a prompt-based annotation protocol, 
(3) a lightweight aggregation mechanism, and 
(4) a bilingual benchmark resource. 
GPT-4o is used as one possible annotation engine rather than a required component of the framework. 
The framework is model-agnostic and can be instantiated with different annotation models under the same developmental taxonomy and prompting protocol. 
Importantly, the taxonomy defines the allowable developmental label space and associated age groups, but does not determine which labels are activated for a given story. 
The open-ended inference step is performed by the annotation model, which identifies the cognitive, linguistic, and social-emotional abilities required for narrative comprehension.

\subsection{Problem Formulation}

Given a narrative text $S$, CLARA derives a developmental label set $\mathcal{L}(S)$, where each label belongs to one of three dimensions:
\[
\mathrm{COG},\ \mathrm{LAN},\ \mathrm{SEL}.
\]

Based on these annotations, CLARA estimates a distribution over developmental age groups:
\[
\mathcal{A} = \{a_1, a_2, \dots, a_K\}.
\]

\subsection{Structured Developmental Annotation}

CLARA uses a predefined developmental taxonomy inspired by developmental psychology and educational theory. 
The taxonomy is hierarchically organized from foundational abilities, such as vocabulary acquisition and emotional expression, to higher-level reasoning abilities including causal reasoning, narrative inference, and perspective taking.

\begin{figure*}[t]
\centering
\includegraphics[width=0.85\textwidth]{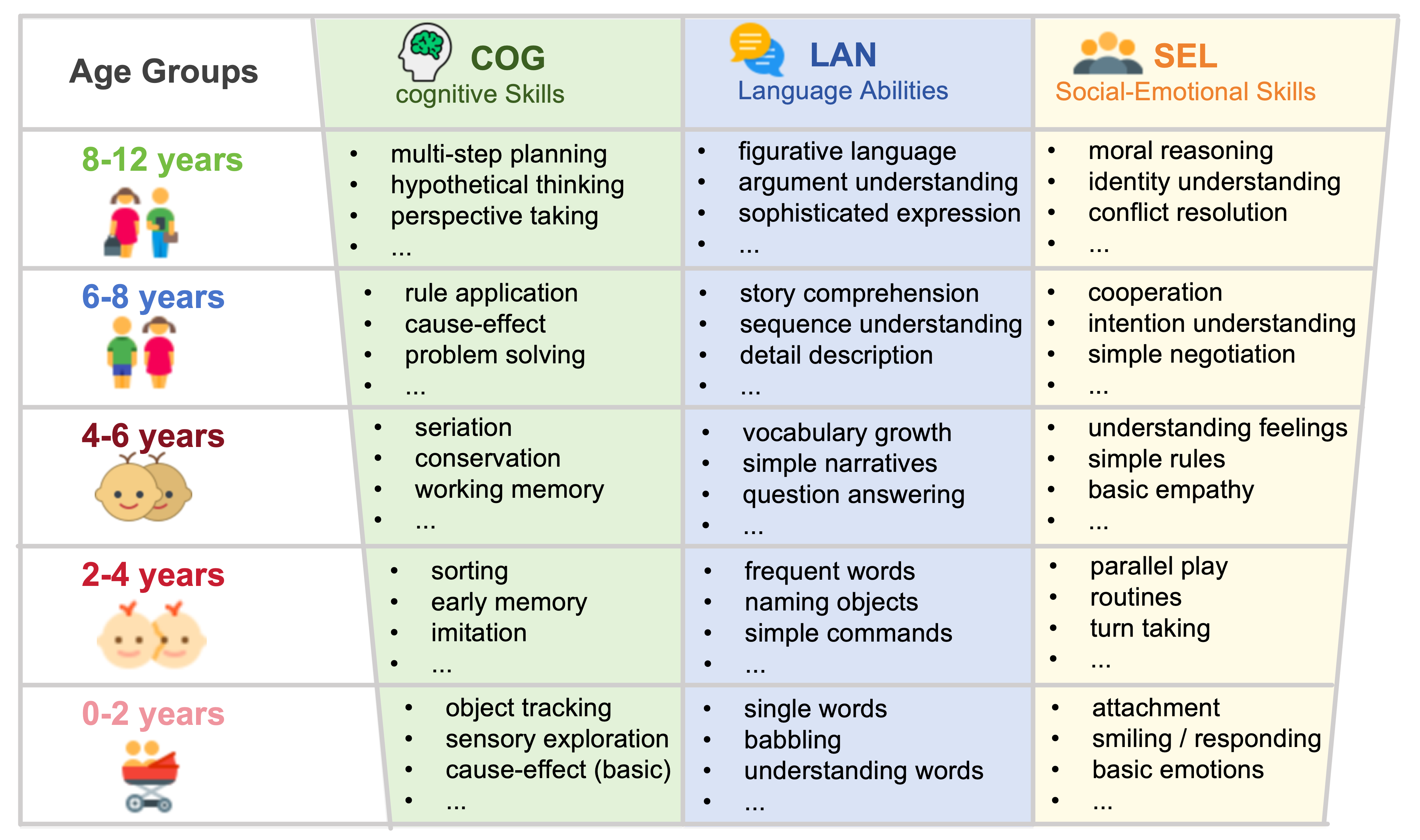}
\caption{Hierarchical developmental annotation framework across COG, LAN, and SEL dimensions.}
\label{fig:funnel}
\end{figure*}

% During annotation, the model is constrained to select labels only from the predefined taxonomy. 
% Each label is associated with one or more developmental age groups and corresponding activation criteria. 
% The complete taxonomy is provided in Appendix~\ref{app:taxonomy}.

During annotation, the model is constrained to select labels only from the predefined taxonomy. 
For each story, the annotation model determines which developmental abilities are required for comprehension and activates the corresponding COG, LAN, and SEL labels. 
Each label is associated with one or more developmental age groups and corresponding activation criteria. 
The complete taxonomy is provided in Appendix~\ref{app:taxonomy}.

\subsection{Developmental Representation Aggregation}

Given the annotated label set $\mathcal{L}(S)$, CLARA first derives an unnormalized developmental score:
\[
\tilde{P}(a_i \mid S)
=
\sum_{l \in \mathcal{L}(S)}
w_l \cdot \mathbf{1}(a_i \in g(l)),
\]
where $g(l)$ denotes the developmental groups associated with label $l$.

The scores are normalized into a developmental distribution:
\[
P(a_i \mid S)
=
\frac{\tilde{P}(a_i \mid S)}
{\sum_{a_j \in \mathcal{A}} \tilde{P}(a_j \mid S)}.
\]

% In the current implementation, all labels are uniformly weighted ($w_l = 1$) for simplicity and interpretability. 
% The competitive performance of the ``LogReg over CLARA Labels'' baseline suggests that the developmental label representation itself contains strong predictive signals. 
% However, CLARA achieves stronger performance without supervised training, indicating that the predefined developmental mapping and aggregation mechanism provide an effective lightweight alternative. 
% We further examine alternative aggregation strategies in Appendix~\ref{app:aggregation_sensitivity}.

% In the current implementation, all labels are uniformly weighted ($w_l = 1$) for simplicity and interpretability. 
% The strong performance of ``LogReg over CLARA Labels'' suggests that the developmental label space itself contains meaningful predictive signals, while CLARA further improves alignment without supervised training. 
% Alternative aggregation strategies are discussed in Appendix~\ref{app:aggregation_sensitivity}.

In the current implementation, all labels are uniformly weighted ($w_l = 1$) for simplicity and interpretability. 
The aggregation step is intentionally kept lightweight and transparent so that the final developmental estimate remains directly traceable to the activated developmental labels, rather than introducing an additional opaque prediction stage. 
The strong performance of ``LogReg over CLARA Labels'' suggests that the developmental label space itself contains meaningful predictive signals, while CLARA further improves alignment without supervised training. 
Alternative aggregation strategies are discussed in Appendix~\ref{app:aggregation_sensitivity}.

The resulting developmental representation is:
\[
\mathbf{P}(S)
=
[P(a_1 \mid S), \dots, P(a_K \mid S)],
\]
and the dominant developmental stage is estimated through:
\[
\hat{a}
=
\arg\max_{a_i \in \mathcal{A}} P(a_i \mid S).
\]

The full distribution $\mathbf{P}(S)$ is the primary developmental representation produced by CLARA; the argmax is used only when a single developmental stage is required for comparison with categorical baselines and reference labels.

\section{Benchmark Resource Construction}
\label{sec:dataset}

To study whether AI systems can approximate human developmental judgments of children's narratives, we construct a bilingual benchmark containing 1107 Chinese--English children's stories collected from storybooks and educational materials in PDF and PPT formats.

Each story includes aligned bilingual narrative text, normalized silver developmental references, and structured annotations across COG, LAN, and SEL dimensions. 
Among them, 1043 stories contain publisher-provided age annotations and are primarily used for quantitative evaluation.

% The processed narratives, annotation prompts, developmental labels, and evaluation splits will be publicly released under the CC BY 4.0 license at:
% \url{https://anonymous.4open.science/r/anonymous_story_dataset}.

We will release annotation prompts, developmental labels, derived annotations, evaluation splits, and metadata under appropriate licenses. Original story text will only be redistributed where legally permitted. The released resources will be available at:
\url{https://anonymous.4open.science/r/anonymous_story_dataset}.

\begin{figure}[t]
\centering
\includegraphics[width=0.98\columnwidth]{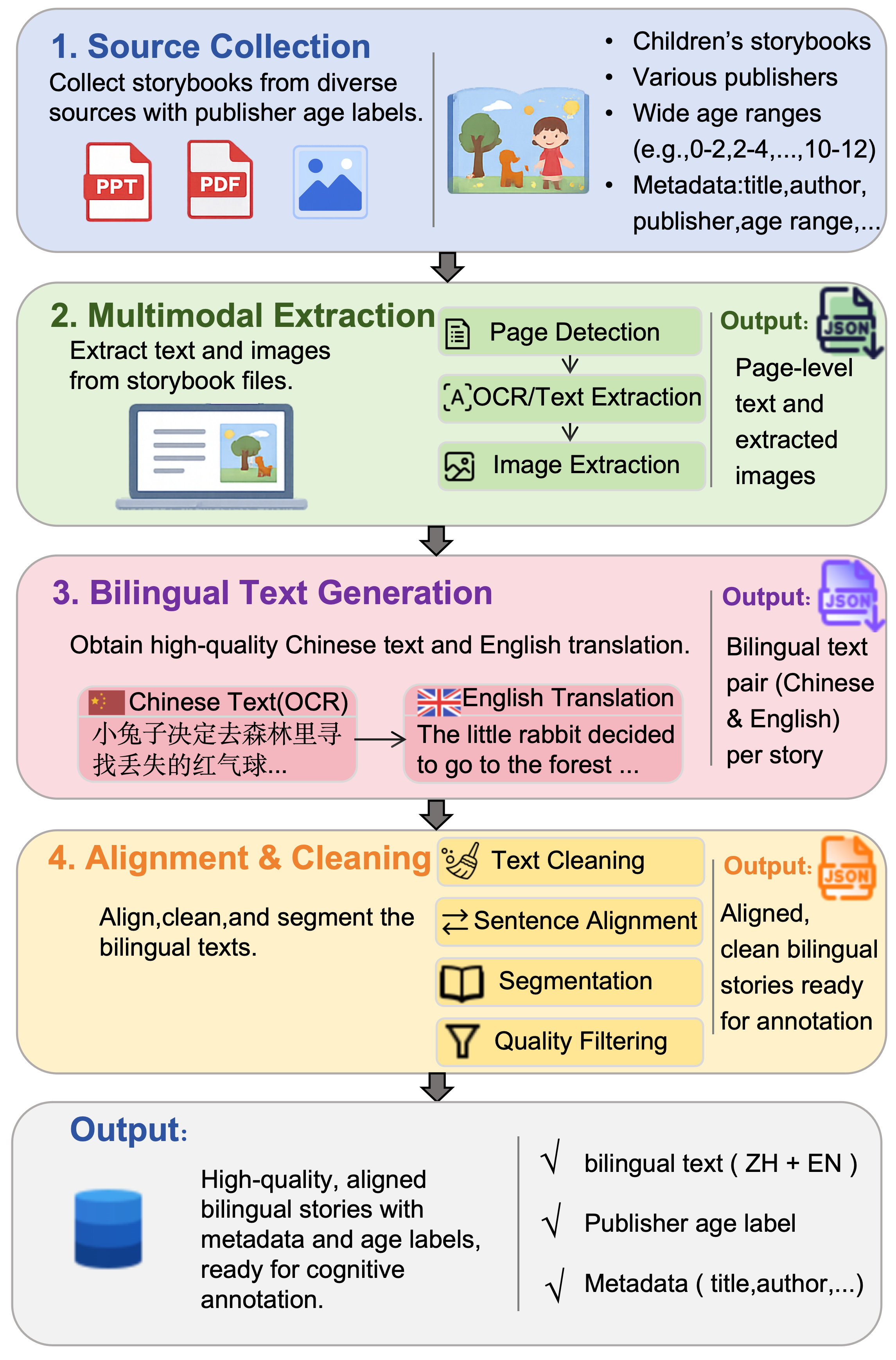}
\caption{Benchmark construction pipeline.}
\label{fig:dataset}
\end{figure}

Figure~\ref{fig:dataset} illustrates the benchmark construction pipeline, including multimodal story extraction, bilingual narrative generation, developmental normalization, and structured annotation.

\subsection{Story Extraction and Data Construction}

The original materials contain both visual and textual content stored in PDF and PPT formats. 
We use Qwen3-VL-32B-Instruct to extract narrative text from storybook pages and convert it into structured Chinese narratives, which are subsequently translated into English to construct aligned bilingual story pairs.

Since individual stories may span multiple pages or slides, related pages are grouped into unified narrative representations during preprocessing. 
Most stories target children between ages 2 and 8 and fall within the medium-length narrative range appropriate for developmental analysis.

\subsection{Data Quality Verification}

Because the benchmark is constructed through multimodal extraction and bilingual translation, we conduct manual quality verification on a randomly sampled subset of 100 stories.

We evaluate extraction correctness, story grouping correctness, and translation adequacy. 
Manual inspection shows that 95\% of sampled stories contain acceptable extraction quality, 93\% preserve correct story grouping, and 91\% maintain adequate bilingual semantic consistency.

\subsection{Silver Developmental References}

Publisher-provided age labels are treated as noisy silver developmental references rather than definitive developmental ground truth. 
We normalize publisher age ranges into five developmental groups:
\[
\{0\text{--}2,\ 2\text{--}4,\ 4\text{--}6,\ 6\text{--}8,\ 8\text{--}12\}.
\]

Broad ranges are assigned to all overlapping developmental groups. 
For example, a story labeled as 5--10 is mapped to 4--6, 6--8, and 8--12.

\begin{table}[t]
\centering
\small
\setlength{\tabcolsep}{6pt}
\begin{tabular}{lcc}
\hline
Age Group & \#Mapped Stories & Share \\
\hline
0--2 & 30 & 2.9\% \\
2--4 & 544 & 52.2\% \\
4--6 & 687 & 65.9\% \\
6--8 & 510 & 48.9\% \\
8--12 & 345 & 33.1\% \\
\hline
Annotated stories & 1043 & 94.2\% \\
Unknown age & 64 & 5.8\% \\
Total stories & 1107 & 100.0\% \\
\hline
\end{tabular}
\caption{Developmental age distribution after normalization.}
\label{tab:age_distribution}
\end{table}

After normalization and filtering ambiguous labels, 1030 stories are retained for quantitative evaluation. 
Benchmark evaluation should therefore be interpreted as measuring alignment with conventional developmental recommendations rather than absolute developmental correctness.

\subsection{Translated Bilingual Consistency}

The bilingual design supports consistency evaluation under translated bilingual settings by comparing developmental predictions and annotations across Chinese and English narrative pairs. Since the English narratives are translations of the corresponding Chinese stories, this analysis evaluates consistency under translation rather than cross-lingual or cross-cultural generalization.

\subsection{Developmental Annotation Analysis}

The taxonomy contains 122 developmental labels distributed across COG, LAN, and SEL dimensions. 
Stories targeting younger children emphasize perceptual abilities and emotional expression, whereas stories for older children increasingly involve causal reasoning, narrative inference, emotional regulation, and perspective taking.

\section{Evaluation and Analysis}
\label{sec:experiments}

We evaluate CLARA from the perspective of developmental narrative understanding through developmental label validation, annotation model comparison, translated bilingual consistency analysis, benchmark comparison, ablation studies, and human evaluation. 
Our evaluation focuses on whether structured developmental annotation can capture meaningful developmental progression patterns, align with human developmental judgments, remain stable across translated bilingual narratives, and provide advantages over readability-based and direct prompting approaches.

\subsection{Evaluation Setup}

The benchmark contains 1107 stories, among which 1030 are retained for quantitative evaluation after normalization and filtering. We additionally construct a balanced 100-story validation subset for annotation model comparison and consistency analysis.

All LLM-based experiments use temperature 0, and annotation model selection is repeated three times. Table~\ref{tab:annotation_model_selection} reports annotation-model comparison over repeated runs on the 100-story validation subset, whereas Table~\ref{tab:benchmark_main} reports the final full-benchmark result using the selected GPT-4o annotation engine. Because the full-benchmark evaluation uses temperature 0 with deterministic downstream aggregation and evaluation, Table~\ref{tab:benchmark_main} reports a single final result. Because publisher age labels are noisy silver references rather than definitive developmental ground truth, the evaluation should be interpreted as measuring alignment with conventional developmental recommendations. Results involving the 0--2 developmental group should be interpreted cautiously because this category contains only 30 stories and developmental suitability at this age may depend more strongly on illustrations, repetition, sensory engagement, and caregiver interaction, which are not modeled in the current text-only formulation.

We evaluate developmental alignment using overlap accuracy and mean developmental distance. For annotation model comparison, we additionally report mapped accuracy, label stability (L-Stab.), and age stability (A-Stab.). Detailed metric definitions are provided in Appendix~\ref{app:metrics}.

\subsection{Empirical Validation of Developmental Labels}

We analyze representative label distributions across developmental stages using annotations generated by GPT-4o, Gemini-2.5-Pro, Qwen-Max, and DeepSeek-Chat.

\begin{figure}[t]
\centering
\includegraphics[width=0.95\columnwidth]{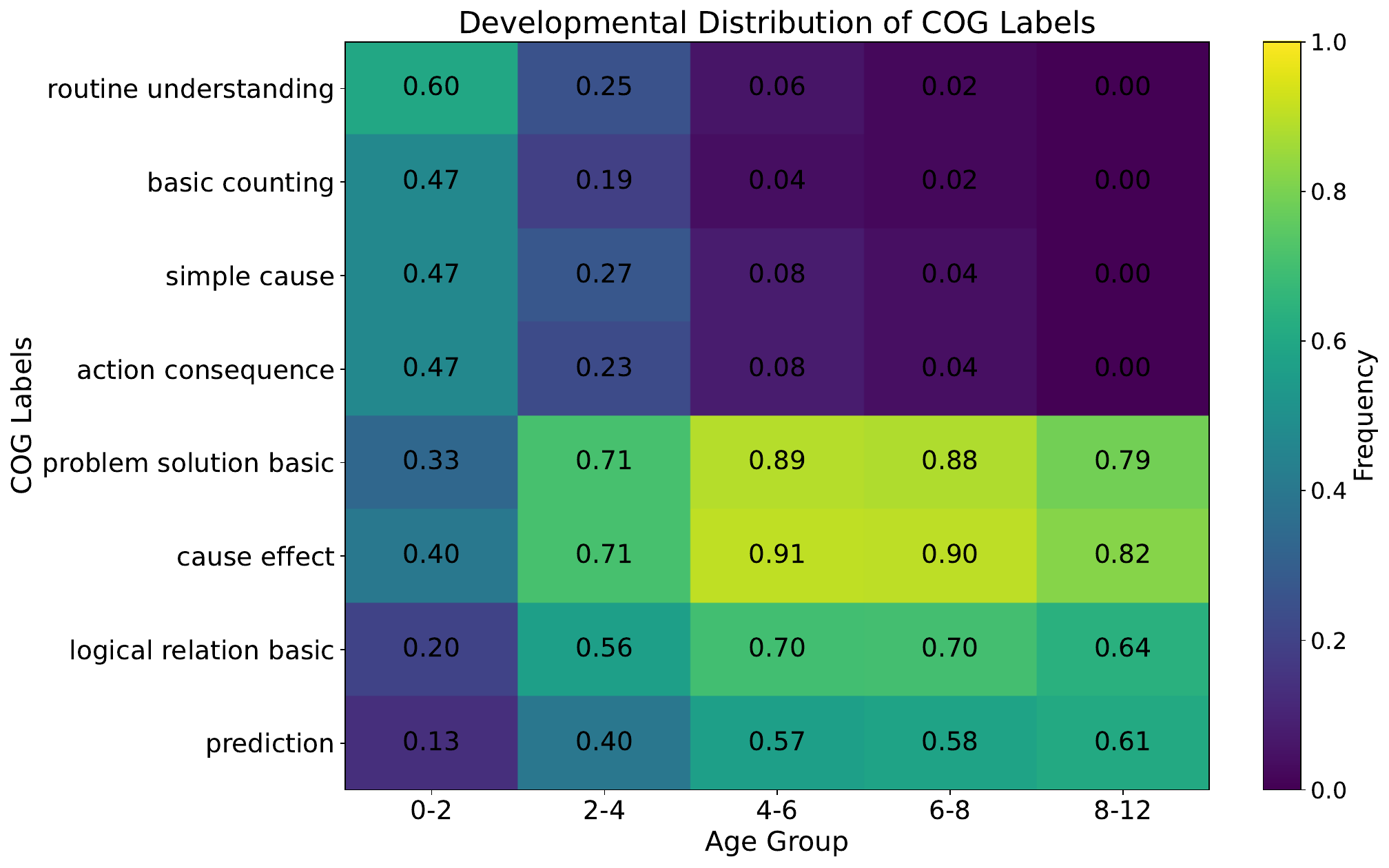}
\caption{Developmental distribution of representative COG labels across age groups.}
\label{fig:cog_heatmap}
\end{figure}

\begin{figure}[t]
\centering
\includegraphics[width=0.95\columnwidth]{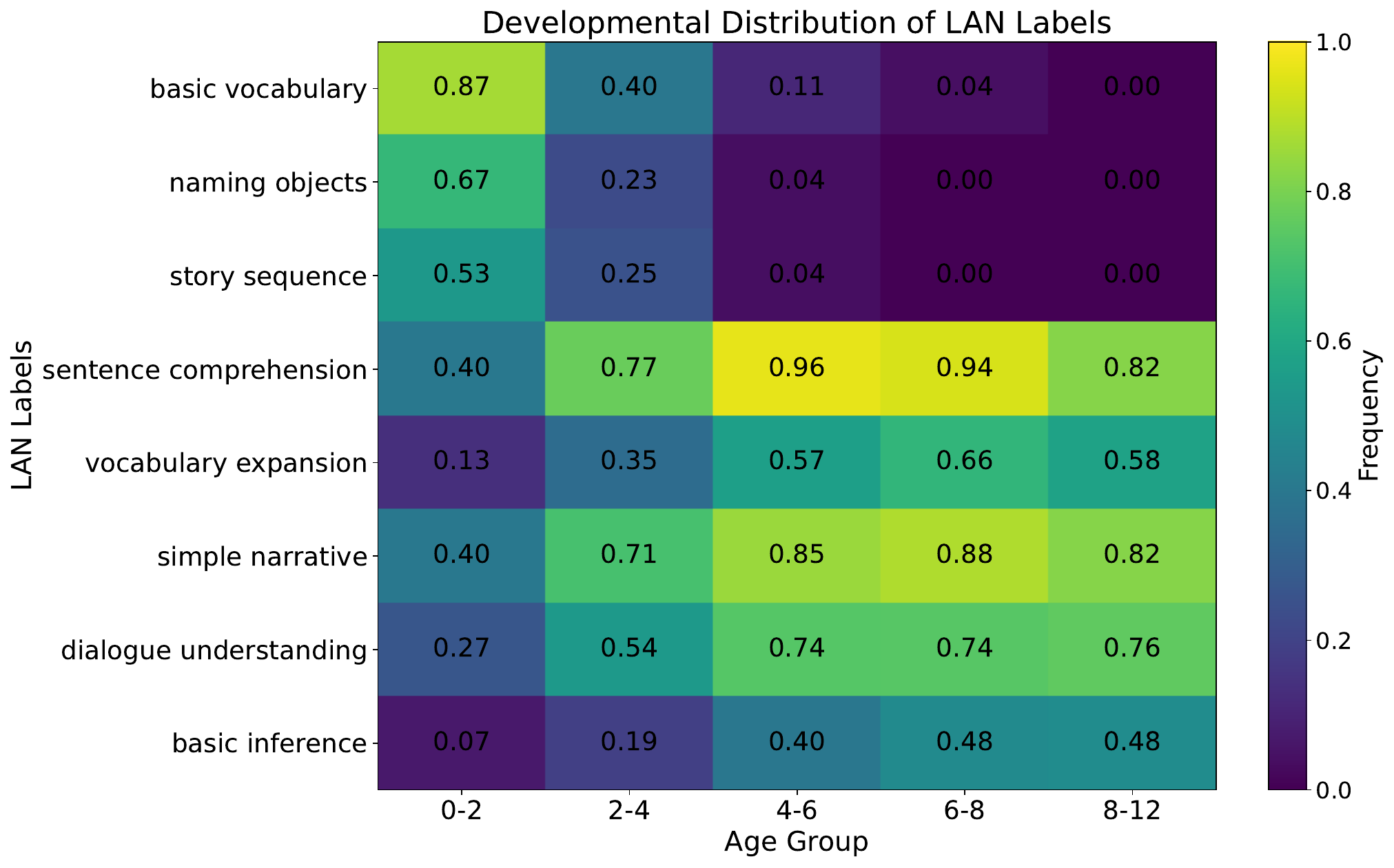}
\caption{Developmental distribution of representative LAN labels across age groups.}
\label{fig:lan_heatmap}
\end{figure}

\begin{figure}[t]
\centering
\includegraphics[width=0.95\columnwidth]{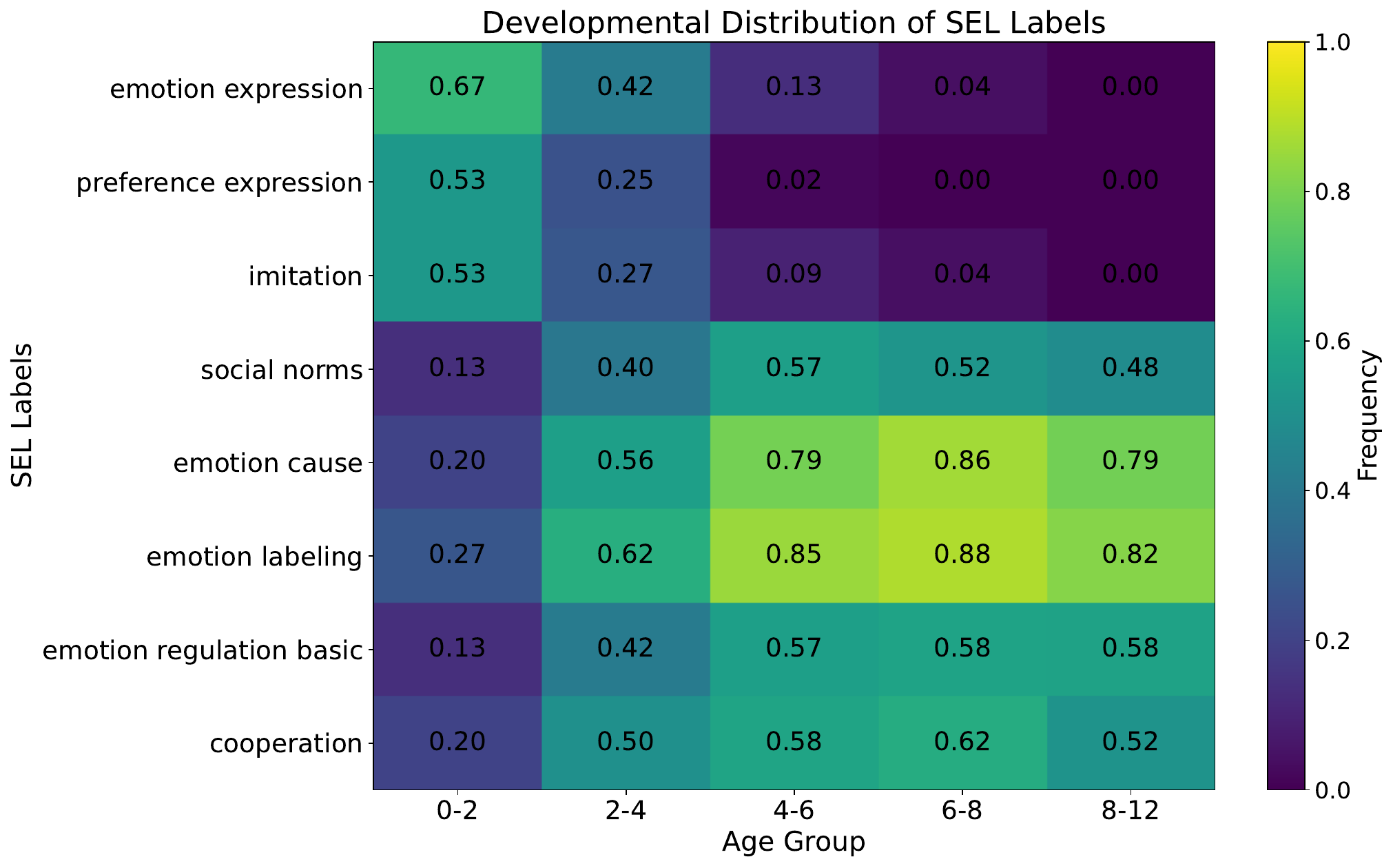}
\caption{Developmental distribution of representative SEL labels across age groups.}
\label{fig:sel_heatmap}
\end{figure}

Figures~\ref{fig:cog_heatmap}, \ref{fig:lan_heatmap}, and \ref{fig:sel_heatmap} show developmental progression trends across COG, LAN, and SEL dimensions. 
Stories targeting younger children emphasize perceptual abilities and emotional expression, whereas stories for older children increasingly involve causal reasoning, narrative inference, emotional regulation, and perspective taking.

These trends remain relatively consistent across annotation models, suggesting that the observed structures capture meaningful developmental signals rather than artifacts of a specific annotation engine. Some foundational labels also remain active in older age groups because developmental abilities are cumulative rather than mutually exclusive. Interpretation of the 0--2 developmental group should nevertheless be treated cautiously due to the substantially smaller number of stories in this category.

\subsection{Annotation Model Comparison}

We compare GPT-4o, Gemini-2.5-Pro, Qwen-Max, and DeepSeek-Chat on the 100-story validation subset using developmental alignment, label coverage, and annotation stability metrics.

\begin{table}[t]
\centering
\footnotesize
\setlength{\tabcolsep}{3pt}
\begin{tabular}{lcccc}
\hline
Model & Acc. & Cov. & L-Stab. & A-Stab. \\
\hline
GPT-4o & 0.71$\pm$0.01 & 17.73$\pm$0.15 & 0.87 & 0.89 \\
Gemini-2.5-Pro & \textbf{0.87}$\pm$0.01 & 29.56$\pm$0.65 & 0.66 & 0.79 \\
Qwen-Max & 0.73$\pm$0.02 & 26.86$\pm$0.71 & 0.77 & 0.79 \\
DeepSeek-Chat & 0.69$\pm$0.02 & 62.41$\pm$0.48 & \textbf{0.91} & \textbf{0.91} \\
\hline
\end{tabular}
\caption{Annotation model comparison on the 100-story validation subset.}
\label{tab:annotation_model_selection}
\end{table}

As shown in Table~\ref{tab:annotation_model_selection}, Gemini-2.5-Pro achieves the highest mapped accuracy but tends to over-label developmental signals, whereas DeepSeek-Chat produces excessively dense annotations. 
GPT-4o achieves the most balanced tradeoff between developmental alignment and annotation stability, and is therefore selected as the primary annotation engine.
Additional annotation validation results on a manually annotated educator subset are provided in Appendix~\ref{app:gold_validation}.

\subsection{Consistency under Translated Bilingual Narratives}

% Additional annotation validation results on a manually annotated educator subset are provided in Appendix~\ref{app:gold_validation}.

To evaluate whether structured developmental representations remain stable under bilingual settings, we compare annotations generated from Chinese and English versions of the same 100 stories using GPT-4o.

\begin{table}[t]
\centering
\small
\setlength{\tabcolsep}{6pt}
\begin{tabular}{lc}
\hline
Metric & Score \\
\hline
Overall Label Jaccard & 0.772 \\
COG Jaccard & 0.793 \\
LAN Jaccard & 0.774 \\
SEL Jaccard & 0.775 \\
Age Agreement & 0.860 \\
Strict Exact Full-Set Match & 0.180 \\
\hline
\end{tabular}
\caption{Consistency under translated bilingual narratives.}
\label{tab:cross_lingual}
\end{table}

As shown in Table~\ref{tab:cross_lingual}, developmental representations remain relatively stable across translated bilingual narratives, with overall annotation similarity of 0.772 and developmental-stage agreement of 86.0\%. 
The relatively low strict exact full-set match reflects the difficulty of obtaining identical fine-grained developmental annotations across translated bilingual narratives. 
However, the substantially higher Jaccard overlap and developmental-stage agreement suggest that higher-level developmental reasoning signals remain comparatively stable despite variation in individual labels.

\subsection{Benchmark Comparison with Readability and Prompting Baselines}
\label{sec:benchmark}

We compare CLARA against readability-based baselines, direct prompting baselines, rubric-guided prompting, and a logistic regression classifier trained over CLARA developmental labels.

% \begin{table}[t]
% \centering
% \small
% \setlength{\tabcolsep}{4pt}
% \begin{tabular}{lcc}
% \hline
% Method & Overlap $\uparrow$ & Mean Distance $\downarrow$ \\
% \hline
% Dale--Chall & 0.469 & 0.882 \\
% FRE & 0.517 & 0.681 \\
% FKGL & 0.603 & 0.502 \\
% GPT-4o Direct & 0.730 & 0.672 \\
% Gemini-2.5-Pro Direct & 0.768 & 0.497 \\
% Qwen-Max Direct & 0.775 & 0.467 \\
% DeepSeek-Chat Direct & 0.787 & 0.451 \\
% GPT-4o Rubric Direct & 0.835 & 0.284 \\
% LogReg over CLARA Labels & 0.874 & 0.173 \\
% CLARA (ours) & \textbf{0.904} & \textbf{0.096} \\
% \hline
% \end{tabular}
% \caption{Benchmark comparison using normalized developmental reference labels.}
% \label{tab:benchmark_main}
% \end{table}

\begin{table}[t]
\centering
\small
\setlength{\tabcolsep}{3pt}
\begin{tabular}{lcc}
\hline
Method & Overlap $\uparrow$ & Mean Distance $\downarrow$ \\
\hline
Dale--Chall & 0.469 & 0.882 \\
FRE & 0.517 & 0.681 \\
FKGL & 0.603 & 0.502 \\
GPT-4o & 0.730 & 0.672 \\
Gemini-2.5-Pro & 0.768 & 0.497 \\
Qwen-Max & 0.775 & 0.467 \\
DeepSeek-Chat & 0.787 & 0.451 \\
GPT-4o + Rubric & 0.835 & 0.284 \\
LogReg + CLARA Labels & 0.874 & 0.173 \\
CLARA (ours) & \textbf{0.904} & \textbf{0.096} \\
\hline
\end{tabular}
\caption{Benchmark comparison using normalized developmental reference labels.}
\label{tab:benchmark_main}
\end{table}

As shown in Table~\ref{tab:benchmark_main}, readability metrics substantially underperform LLM-based approaches, suggesting that surface linguistic complexity alone is insufficient for developmental narrative understanding. The rubric-guided baseline improves over direct prompting, indicating that intermediate reasoning contributes to developmental estimation, while the strong performance of ``LogReg over CLARA Labels'' further shows that the structured developmental annotations themselves contain meaningful developmental signals. CLARA nevertheless achieves the strongest alignment with normalized developmental references, suggesting that explicit developmental annotation and structured aggregation provide additional benefits beyond implicit reasoning alone.

\subsection{Contribution Decomposition}
\label{sec:decomposition}

To better understand which components contribute most to CLARA's performance, we conduct a contribution decomposition study.

% \begin{table}[b]
% \centering
% \small
% \setlength{\tabcolsep}{4pt}
% \begin{tabular}{lcc}
% \hline
% Variant & Overlap $\uparrow$ & Mean Distance $\downarrow$ \\
% \hline
% GPT-4o Direct & 0.730 & 0.672 \\
% Free-form Rationale + Aggregation & 0.791 & 0.411 \\
% Taxonomy Labels Only & 0.852 & 0.231 \\
% Taxonomy + Age Mapping & 0.889 & 0.148 \\
% Full CLARA & \textbf{0.904} & \textbf{0.096} \\
% \hline
% \end{tabular}
% \caption{Contribution decomposition analysis for CLARA.}
% \label{tab:decomposition}
% \end{table}

\begin{table}[b]
\centering
\small
\setlength{\tabcolsep}{5pt}
\begin{tabular}{lcc}
\hline
Variant & Overlap $\uparrow$ & Mean Distance $\downarrow$ \\
\hline
GPT-4o Direct & 0.730 & 0.672 \\
Free-form + Agg. & 0.791 & 0.411 \\
Taxonomy Labels & 0.852 & 0.231 \\
Taxonomy + Mapping & 0.889 & 0.148 \\
Full CLARA & \textbf{0.904} & \textbf{0.096} \\
\hline
\end{tabular}
\caption{Contribution decomposition analysis for CLARA.}
\label{tab:decomposition}
\end{table}

Table~\ref{tab:decomposition} shows progressive improvement as additional structured components are introduced. 
The taxonomy itself contributes substantially to developmental alignment, while explicit developmental age mapping and structured aggregation further improve performance. 
These results suggest that CLARA's gains do not originate solely from prompt engineering, but from the interaction between structured developmental representation, constrained annotation, and developmental aggregation.

\subsection{Taxonomy Quality Ablation}

To evaluate the contribution of the developmental taxonomy, we construct several degraded taxonomy variants.

\begin{table}[t]
\centering
\small
\setlength{\tabcolsep}{4pt}
\begin{tabular}{lcc}
\hline
Variant & Overlap $\uparrow$ & Distance $\downarrow$ \\
\hline
Random Mapping & 0.611 & 0.593 \\
Dimension-swapped & 0.742 & 0.366 \\
Coarse Taxonomy & 0.861 & 0.181 \\
Full Taxonomy & \textbf{0.904} & \textbf{0.096} \\
\hline
\end{tabular}
\caption{Taxonomy quality ablation study.}
\label{tab:taxonomy_ablation}
\end{table}

As shown in Table~\ref{tab:taxonomy_ablation}, degraded taxonomies substantially reduce developmental alignment quality, suggesting that CLARA benefits from meaningful developmental organization rather than prompt structure alone. Additional dimension-level ablation results are provided in Appendix~\ref{app:dimension_ablation}.

\subsection{Human Evaluation}
\label{sec:human_eval}

We conduct a blinded human evaluation study to assess whether the proposed developmental annotations align with educational judgments from early-childhood educators. We recruit eight educators with early-childhood education experience and randomly sample 60 stories covering multiple developmental stages from the benchmark, where each story is independently evaluated by three educators, resulting in 180 evaluation instances.

\begin{table}[t]
\centering
\small
\setlength{\tabcolsep}{5pt}
\begin{tabular}{lc}
\hline
Criterion & Score \\
\hline
COG Label Quality & 4.24 \\
LAN Label Quality & 4.18 \\
SEL Label Quality & 4.31 \\
Educational Interpretability & \textbf{4.42} \\
Predicted Age Appropriateness & 4.17 \\
\hline
Average & 4.26 \\
\hline
\end{tabular}
\caption{Blinded human evaluation results.}
\label{tab:human_eval}
\end{table}

As shown in Table~\ref{tab:human_eval}, educational interpretability receives the highest average score, suggesting that the proposed annotation space provides meaningful developmental explanations beyond direct age prediction alone. Educators further noted that structured developmental labels better expose emotional reasoning, social interaction, and implicit causal understanding, while the average pairwise annotation agreement across the three evaluators per story reaches 0.81, supporting the reliability of the proposed developmental annotation framework.

\subsection{Human Developmental Ranking Validation}
\label{sec:human_ranking}

To provide a more independent evaluation beyond publisher-provided developmental references, we conduct a blinded human developmental ranking study using 60 story pairs covering both adjacent and distant developmental stages, and compare human pairwise rankings against readability metrics, direct prompting baselines, rubric-guided prompting, and CLARA developmental representations.

\begin{table}[t]
\centering
\small
\setlength{\tabcolsep}{4pt}
\begin{tabular}{lc}
\hline
Method & Human Agreement \\
\hline
FKGL & 0.58 \\
GPT-4o Direct & 0.67 \\
GPT-4o Rubric Direct & 0.73 \\
CLARA (ours) & \textbf{0.81} \\
\hline
\end{tabular}
\caption{Agreement with blinded human developmental ranking judgments.}
\label{tab:human_ranking}
\end{table}

As shown in Table~\ref{tab:human_ranking}, CLARA achieves the highest agreement with human developmental judgments, suggesting that structured developmental annotation better captures relative developmental complexity than readability-based or direct prompting approaches.

\subsection{Analysis of Developmental Label Discrepancies}
\label{sec:age_discrepancy}

We analyze disagreement cases between publisher developmental references and CLARA developmental estimates using 30 randomly sampled stories evaluated by blinded educators.

\begin{table}[t]
\centering
\small
\setlength{\tabcolsep}{4pt}
\begin{tabular}{lc}
\hline
Preference & Human Evaluators \\
\hline
Prefer CLARA & 61\% \\
Prefer Publisher & 19\% \\
Both Reasonable & 20\% \\
\hline
\end{tabular}
\caption{Human evaluation on disagreement cases between publisher developmental references and CLARA developmental estimates.}
\label{tab:age_discrepancy}
\end{table}

% As shown in Table~\ref{tab:age_discrepancy}, educators more frequently prefer CLARA developmental estimates over publisher recommendations. 
% Many disagreement cases involve abstract reasoning, emotional nuance, or social understanding that publisher labels may underrepresent.

As shown in Table~\ref{tab:age_discrepancy}, educators more frequently prefer CLARA developmental estimates over publisher recommendations. 
Many disagreement cases involve abstract reasoning, emotional nuance, or social understanding that publisher labels may underrepresent. 
Disagreements may arise from several sources, including broad publisher recommendations, stories containing mixed developmental signals across adjacent age bands, and occasional over- or under-activation of developmental labels by the annotation model.

\section{Conclusion}

In this paper, we study whether AI systems can approximate human developmental judgments of children’s narratives through structured developmental annotation. We introduce CLARA, a cognitively grounded approach for developmental narrative understanding across COG, LAN, and SEL dimensions, together with a translated bilingual benchmark containing 1107 Chinese--English children’s stories and normalized silver developmental references. Experimental results show that structured developmental annotation achieves substantially stronger alignment with developmental references and blinded human educational judgments than readability-based methods and direct prompting baselines, suggesting that developmental narrative understanding cannot be fully captured through surface linguistic complexity alone. Additional analyses further demonstrate relatively stable developmental representations across translated bilingual narratives and annotation models. Overall, our findings indicate that AI systems can approximate certain aspects of human developmental reasoning when guided by interpretable developmental representations, while also highlighting the importance of transparency, human oversight, and cognitively grounded evaluation in educational NLP.

\section*{Limitations}

Despite promising results, this work has several limitations. Developmental narrative understanding is inherently subjective and may vary across educational, cultural, and linguistic contexts. The proposed framework relies on structured developmental annotations and silver developmental references, which may not fully capture the complexity and diversity of human developmental reasoning. In addition, the current study mainly focuses on text-based narrative analysis and does not explicitly model multimodal storytelling elements such as illustrations, visual narrative structure, repetition, or caregiver interaction. This limitation is particularly important for the 0--2 age group, which contains only 30 stories and may rely more heavily on non-textual developmental cues. Moreover, because the English narratives are translations of the corresponding Chinese stories, the bilingual analysis should be interpreted as consistency under translation rather than evidence of broad cross-lingual or cross-cultural generalization. Future work may further explore more diverse developmental settings, multimodal evaluation, and stronger reasoning-oriented approaches for developmental narrative understanding.

\section*{Ethics Statement}

This work studies developmental narrative understanding for educational and research purposes. The proposed framework is intended to support interpretable analysis of children’s narratives rather than automated educational judgment, psychological assessment, or clinical decision-making. Because developmental suitability is inherently influenced by cultural, linguistic, educational, and individual factors, the developmental annotations and predictions produced by CLARA should be interpreted as supportive computational signals rather than authoritative developmental conclusions. In addition, the framework relies on large language model annotations and silver developmental references, which may introduce biases and inconsistencies inherited from pretrained models, prompting strategies, and publisher conventions. The story corpus consists of published children’s narratives and educational materials rather than personal-user data, and no personally identifying information is included in the released benchmark. Human evaluation was conducted anonymously with voluntary participation from educators, and no personal or sensitive information was collected. Participants were recruited through direct invitation based on their early-childhood education experience and participated voluntarily without monetary compensation. All participants provided informed consent prior to participation and were informed that their anonymized responses would be used for research purposes. Overall, CLARA is intended as a research framework for studying developmental narrative understanding and interpretable educational NLP.

\section*{Acknowledgments}

We thank the anonymous reviewers, educators, and evaluators for their valuable feedback and developmental perspectives on children's narrative understanding, as well as all contributors who provided discussions, annotation support, and technical assistance throughout this project. Code, prompts, derived annotations, metadata, and permitted benchmark components will be released upon acceptance to support future research in developmental narrative understanding and educational NLP.

\bibliography{custom_short}

@inproceedings{mostafazadeh-etal-2016-corpus,
    title = "A Corpus and Cloze Evaluation for Deeper Understanding of Commonsense Stories",
    author = "Mostafazadeh, Nasrin  and
      Chambers, Nathanael  and
      He, Xiaodong  and
      Parikh, Devi  and
      Batra, Dhruv  and
      Vanderwende, Lucy  and
      Kohli, Pushmeet  and
      Allen, James",
    editor = "Knight, Kevin  and
      Nenkova, Ani  and
      Rambow, Owen",
    booktitle = "Proceedings of the 2016 Conference of the North {A}merican Chapter of the Association for Computational Linguistics: Human Language Technologies",
    month = jun,
    year = "2016",
    address = "San Diego, California",
    publisher = "Association for Computational Linguistics",
    url = "https://aclanthology.org/N16-1098/",
    doi = "10.18653/v1/N16-1098",
    pages = "839--849"
}

@article{flesch1948new,
  title = {A New Readability Yardstick},
  author = {Flesch, Rudolf},
  journal = {Journal of Applied Psychology},
  volume = {32},
  number = {3},
  pages = {221--233},
  year = {1948},
  doi = {10.1037/h0057532}
}

@techreport{kincaid1975derivation,
  title = {Derivation of New Readability Formulas (Automated Readability Index, Fog Count and Flesch Reading Ease Formula) for Navy Enlisted Personnel},
  author = {Kincaid, J. Peter and Fishburne, Robert P. Jr. and Rogers, Richard L. and Chissom, Brad S.},
  institution = {Naval Technical Training Command},
  address = {Millington, TN, USA},
  number = {Research Branch Report 8-75},
  year = {1975},
  url = {https://stars.library.ucf.edu/istlibrary/56/}
}

@article{martinc2021readability,
  title = {Supervised and Unsupervised Neural Approaches to Text Readability},
  author = {Martinc, Matej and Pollak, Senja and Robnik-{\v{S}}ikonja, Marko},
  journal = {Computational Linguistics},
  volume = {47},
  number = {1},
  pages = {141--179},
  year = {2021},
  doi = {10.1162/coli_a_00398},
  url = {https://direct.mit.edu/coli/article/47/1/141/97338/Supervised-and-Unsupervised-Neural}
}

@book{piaget1952origins,
  title = {The Origins of Intelligence in Children},
  author = {Piaget, Jean},
  publisher = {International Universities Press},
  address = {New York, NY},
  year = {1952},
  url = {https://books.google.com/books/about/The_Origins_of_Intelligence_in_Children.html?id=fKa9AAAAMAAJ}
}

@book{vygotsky1978mind,
  title = {Mind in Society: The Development of Higher Psychological Processes},
  author = {Vygotsky, Lev S.},
  publisher = {Harvard University Press},
  address = {Cambridge, MA},
  year = {1978},
  isbn = {9780674576292},
  url = {https://www.hup.harvard.edu/books/9780674576292}
}

@techreport{casel2012framework,
  title = {2013 {CASEL} Guide: Effective Social and Emotional Learning Programs---Preschool and Elementary School Edition},
  author = {{Collaborative for Academic, Social, and Emotional Learning}},
  institution = {Collaborative for Academic, Social, and Emotional Learning},
  address = {Chicago, IL},
  year = {2012},
  url = {https://files.eric.ed.gov/fulltext/ED581699.pdf}
}

@inproceedings{wang2022self,
  title = {Self-Consistency Improves Chain of Thought Reasoning in Language Models},
  author = {Wang, Xuezhi and Wei, Jason and Schuurmans, Dale and Le, Quoc V. and Chi, Ed H. and Narang, Sharan and Chowdhery, Aakanksha and Zhou, Denny},
  booktitle = {International Conference on Learning Representations},
  year = {2023},
  url = {https://openreview.net/forum?id=1PL1NIMMrw}
}

@inproceedings{wei2022chain,
  title = {Chain-of-Thought Prompting Elicits Reasoning in Large Language Models},
  author = {Wei, Jason and Wang, Xuezhi and Schuurmans, Dale and Bosma, Maarten and Ichter, Brian and Xia, Fei and Chi, Ed H. and Le, Quoc V. and Zhou, Denny},
  booktitle = {Advances in Neural Information Processing Systems},
  volume = {35},
  pages = {24824--24837},
  publisher = {Curran Associates, Inc.},
  year = {2022},
  url = {https://proceedings.neurips.cc/paper_files/paper/2022/hash/9d5609613524ecf4f15af0f7b31abca4-Abstract-Conference.html}
}

@misc{zhou2023least,
      title={Least-to-Most Prompting Enables Complex Reasoning in Large Language Models}, 
      author={Denny Zhou and Nathanael Schärli and Le Hou and Jason Wei and Nathan Scales and Xuezhi Wang and Dale Schuurmans and Claire Cui and Olivier Bousquet and Quoc Le and Ed Chi},
      year={2023},
      eprint={2205.10625},
      archivePrefix={arXiv},
      primaryClass={cs.AI},
      url={https://arxiv.org/abs/2205.10625}, 
}

@article{gilardi2023chatgpt,
  title = {ChatGPT Outperforms Crowd Workers for Text-Annotation Tasks},
  author = {Gilardi, Fabrizio and Alizadeh, Meysam and Kubli, Ma{\"e}l},
  journal = {Proceedings of the National Academy of Sciences},
  volume = {120},
  number = {30},
  pages = {e2305016120},
  year = {2023},
  doi = {10.1073/pnas.2305016120},
  url = {https://www.pnas.org/doi/10.1073/pnas.2305016120}
}

@article{dale1948formula,
  title = {A Formula for Predicting Readability},
  author = {Dale, Edgar and Chall, Jeanne S.},
  journal = {Educational Research Bulletin},
  volume = {27},
  number = {1},
  pages = {11--20, 28},
  year = {1948},
  url = {https://www.jstor.org/stable/1473169}
}

@inproceedings{sap2019atomic,
  title = {{ATOMIC}: An Atlas of Machine Commonsense for If-Then Reasoning},
  author = {Sap, Maarten and Le Bras, Ronan and Allaway, Emily and Bhagavatula, Chandra and Lourie, Nicholas and Rashkin, Hannah and Roof, Brendan and Smith, Noah A. and Choi, Yejin},
  booktitle = {Proceedings of the AAAI Conference on Artificial Intelligence},
  volume = {33},
  pages = {3027--3035},
  publisher = {AAAI Press},
  year = {2019},
  doi = {10.1609/aaai.v33i01.33013027},
  url = {https://ojs.aaai.org/index.php/AAAI/article/view/4160}
}

@inproceedings{xu2022fairytaleqa,
    title = "Fantastic Questions and Where to Find Them: {F}airytale{QA} {--} An Authentic Dataset for Narrative Comprehension",
    author = "Xu, Ying  and
      Wang, Dakuo  and
      Yu, Mo  and
      Ritchie, Daniel  and
      Yao, Bingsheng  and
      Wu, Tongshuang  and
      Zhang, Zheng  and
      Li, Toby Jia-Jun  and
      Bradford, Nora  and
      Sun, Branda  and
      Hoang, Tran Bao  and
      Sang, Yisi  and
      Hou, Yufang  and
      Ma, Xiaojuan  and
      Yang, Diyi  and
      Peng, Nanyun  and
      Yu, Zhou  and
      Warschauer, Mark",
    editor = "Muresan, Smaranda  and
      Nakov, Preslav  and
      Villavicencio, Aline",
    booktitle = "Proceedings of the 60th Annual Meeting of the Association for Computational Linguistics (Volume 1: Long Papers)",
    month = may,
    year = "2022",
    address = "Dublin, Ireland",
    publisher = "Association for Computational Linguistics",
    url = "https://aclanthology.org/2022.acl-long.34/",
    doi = "10.18653/v1/2022.acl-long.34",
    pages = "447--460"
}

@book{astington1988developing,
  title = {Developing Theories of Mind},
  editor = {Astington, Janet W. and Harris, Paul L. and Olson, David R.},
  address = {Cambridge},
  publisher = {Cambridge University Press},
  year = {1988},
  url = {https://www.finna.fi/Record/luc.991664783806246}
}

@article{wellman2004scaling,
  title = {Scaling of Theory-of-Mind Tasks},
  author = {Wellman, Henry M. and Liu, David},
  journal = {Cognition},
  volume = {92},
  number = {3},
  pages = {167--190},
  year = {2004},
  doi = {10.1016/j.cognition.2003.10.004},
  url = {https://www.sciencedirect.com/science/article/pii/S0010027703002241}
}

@article{gagarina2019main,
  title = {{MAIN}: Multilingual Assessment Instrument for Narratives -- Revised},
  author = {Gagarina, Natalia and Klop, Daleen and Kunnari, Sari and Tantele, Koula and V{\"a}limaa, Taina and Bohnacker, Ute and Walters, Joel},
  journal = {ZAS Papers in Linguistics},
  volume = {63},
  pages = {20},
  year = {2019},
  month = {Nov.},
  doi = {10.21248/zaspil.63.2019.516},
  url = {https://zaspil.leibniz-zas.de/article/view/516}
}

@article{gagarina2022referentiality,
  title = {A New Perspective on Referentiality in Elicited Narratives: Introduction to the Special Issue},
  author = {Gagarina, Natalia and Bohnacker, Ute},
  journal = {First Language},
  volume = {42},
  number = {2},
  pages = {171--190},
  year = {2022},
  doi = {10.1177/01427237211062300},
  url = {https://journals.sagepub.com/doi/10.1177/01427237211062300}
}

@article{kupersmitt2019causal,
  title = {The Linguistic Expression of Causal Relations in Picture-Based Narratives: A Comparative Study of Bilingual and Monolingual Children with TLD and DLD},
  author = {Kupersmitt, Judy R. and Armon-Lotem, Sharon},
  journal = {First Language},
  volume = {39},
  number = {3},
  pages = {319--343},
  year = {2019},
  doi = {10.1177/0142723719831927},
  url = {https://journals.sagepub.com/doi/10.1177/0142723719831927}
}

@inproceedings{vandijk2023theory,
    title = "Theory of Mind in Freely-Told Children{'}s Narratives: A Classification Approach",
    author = "van Dijk, Bram  and
      Spruit, Marco  and
      van Duijn, Max",
    editor = "Rogers, Anna  and
      Boyd-Graber, Jordan  and
      Okazaki, Naoaki",
    booktitle = "Findings of the Association for Computational Linguistics: ACL 2023",
    month = jul,
    year = "2023",
    address = "Toronto, Canada",
    publisher = "Association for Computational Linguistics",
    url = "https://aclanthology.org/2023.findings-acl.822/",
    doi = "10.18653/v1/2023.findings-acl.822",
    pages = "12979--12993"
}

@inproceedings{mostafazadeh2020glucose,
  title = {{GLUCOSE}: GeneraLized and COntextualized Story Explanations},
  author = {Mostafazadeh, Nasrin and Kalyanpur, Aditya and Moon, Lori and Buchanan, David and Berkowitz, Lauren and Biran, Or and Chu-Carroll, Jennifer},
  booktitle = {Proceedings of the 2020 Conference on Empirical Methods in Natural Language Processing (EMNLP)},
  pages = {4569--4586},
  address = {Online},
  publisher = {Association for Computational Linguistics},
  year = {2020},
  doi = {10.18653/v1/2020.emnlp-main.370},
  url = {https://aclanthology.org/2020.emnlp-main.370/}
}

@inproceedings{tikhonov2021storydb,
  title = {{StoryDB}: Broad Multi-language Narrative Dataset},
  author = {Tikhonov, Alexey and Samenko, Igor and Yamshchikov, Ivan P.},
  booktitle = {Proceedings of the 2nd Workshop on Evaluation and Comparison of NLP Systems},
  pages = {32--39},
  address = {Punta Cana, Dominican Republic},
  publisher = {Association for Computational Linguistics},
  year = {2021},
  doi = {10.18653/v1/2021.eval4nlp-1.4},
  url = {https://aclanthology.org/2021.eval4nlp-1.4/}
}

@inproceedings{rashkin2018naive,
  title = {Modeling Naive Psychology of Characters in Simple Commonsense Stories},
  author = {Rashkin, Hannah and Bosselut, Antoine and Sap, Maarten and Knight, Kevin and Choi, Yejin},
  booktitle = {Proceedings of the 56th Annual Meeting of the Association for Computational Linguistics (Volume 1: Long Papers)},
  pages = {2289--2299},
  address = {Melbourne, Australia},
  publisher = {Association for Computational Linguistics},
  year = {2018},
  doi = {10.18653/v1/P18-1213},
  url = {https://aclanthology.org/P18-1213/}
}

@article{reese2011coherence,
  title = {Coherence of Personal Narratives Across the Lifespan: A Multidimensional Model and Coding Method},
  author = {Reese, Elaine and Haden, Catherine A. and Baker-Ward, Lynne and Bauer, Patricia J. and Fivush, Robyn and Ornstein, Peter A.},
  journal = {Journal of Cognition and Development},
  volume = {12},
  number = {4},
  pages = {424--462},
  year = {2011},
  doi = {10.1080/15248372.2011.587854},
  url = {https://doi.org/10.1080/15248372.2011.587854}
}

@article{yang2022bilingual,
  title = {Bilingual Mandarin-English Preschoolers' Spoken Narrative Skills and Contributing Factors: A Remote Online Story-Retell Study},
  author = {Yang, Jingdan and Kim, Jae-Hyun and Tuomainen, Outi and Xu Rattanasone, Nan},
  journal = {Frontiers in Psychology},
  volume = {13},
  pages = {797602},
  year = {2022},
  doi = {10.3389/fpsyg.2022.797602},
  url = {https://www.frontiersin.org/journals/psychology/articles/10.3389/fpsyg.2022.797602}
}

\appendix

\section{AI Usage Statement}

AI tools were used only for literature search assistance, data processing support, and annotation generation during the research process. The writing, organization, interpretation, and final wording of the manuscript were completed and reviewed by the authors. The authors take full responsibility for the content of this paper.

\section{Detailed Evaluation Metrics}
\label{app:metrics}

We evaluate developmental alignment using overlap accuracy and mean developmental distance.

Given predicted developmental groups $\hat{\mathcal{A}}$ and silver reference groups $\mathcal{A}^{*}$, overlap accuracy is defined as:

\[
\mathrm{Overlap}
=
\frac{
|\hat{\mathcal{A}} \cap \mathcal{A}^{*}|
}{
|\hat{\mathcal{A}} \cup \mathcal{A}^{*}|
}.
\]

Mean developmental distance measures the average ordinal distance between predicted and reference developmental groups after mapping developmental stages into ordered indices.

% For annotation model comparison, mapped accuracy measures whether the dominant predicted developmental group overlaps with the normalized silver developmental references.

For annotation model comparison, mapped accuracy measures whether the dominant predicted developmental group overlaps with any of the normalized silver reference groups.

Label stability (L-Stab.) is computed using Jaccard similarity between repeated annotation runs:

% \[
% \mathrm{L\mbox{-}Stab}
% =
% \frac{
% |L_1 \cap L_2|
% }{
% |L_1 \cup L_2|
% },
% \]

\[
\mathrm{L\mbox{-}Stab}
=
\frac{
|\mathcal{L}_1 \cap \mathcal{L}_2|
}{
|\mathcal{L}_1 \cup \mathcal{L}_2|
},
\]

% where $L_1$ and $L_2$ denote developmental label sets generated from repeated runs.
where $\mathcal{L}_1$ and $\mathcal{L}_2$ denote developmental label sets generated from repeated runs.

Age stability (A-Stab.) measures agreement between dominant developmental-stage predictions across repeated annotation runs.

\section{Aggregation Sensitivity Analysis}
\label{app:aggregation_sensitivity}

CLARA currently adopts a uniform weighting strategy in which all developmental labels contribute equally during developmental aggregation. 
This design is intentionally lightweight and interpretable, allowing the framework to focus on structured developmental representations rather than manually engineered weighting heuristics.

To examine whether the framework is overly sensitive to the specific aggregation strategy, we additionally explored several alternative weighting schemes, including frequency-normalized weighting and dimension-balanced weighting across COG, LAN, and SEL dimensions.

Preliminary experiments showed that alternative weighting strategies produced relatively similar developmental alignment trends, while also introducing additional hyperparameters and reduced interpretability. 
We therefore adopt uniform weighting in the main experiments for simplicity, transparency, and reproducibility.

These observations further suggest that the primary contribution of CLARA lies in the structured developmental annotation space itself rather than a particular aggregation heuristic.

\section{Dimension-Level Ablation Study}
\label{app:dimension_ablation}

To further analyze the contribution of different developmental dimensions in CLARA, we conduct a dimension-level ablation study by removing one developmental dimension at a time, including COG, LAN, and SEL.

\begin{table}[t]
\centering
\small
\setlength{\tabcolsep}{5pt}
\begin{tabular}{lcc}
\hline
Variant & Overlap $\uparrow$ & Mean Distance $\downarrow$ \\
\hline
Full CLARA & \textbf{0.904} & \textbf{0.096} \\
w/o COG & 0.861 & 0.164 \\
w/o LAN & 0.873 & 0.151 \\
w/o SEL & 0.842 & 0.183 \\
\hline
\end{tabular}
\caption{Dimension-level ablation study on different developmental dimensions in CLARA.}
\label{tab:dimension_ablation}
\end{table}

Table~\ref{tab:dimension_ablation} shows that removing any developmental dimension reduces developmental alignment and increases developmental distance. Among the three variants, removing SEL leads to the largest performance degradation, suggesting that social-emotional development provides important signals beyond linguistic and cognitive complexity alone. Removing COG and LAN also negatively affects performance, indicating that developmental reasoning benefits from integrating complementary developmental dimensions.

\section{Gold Annotation Validation}
\label{app:gold_validation}

To directly evaluate the developmental annotations, we construct a manually annotated gold subset for label-level comparison.

We randomly sample 60 stories from the benchmark. Three educators from the evaluation pool independently annotate developmental dimensions and representative developmental labels for each story.

\begin{table}[t]
\centering
\small
\setlength{\tabcolsep}{5pt}
\begin{tabular}{lc}
\hline
Metric & Score \\
\hline
COG F1 & 0.84 \\
LAN F1 & 0.81 \\
SEL F1 & 0.79 \\
Representative Macro F1 & 0.73 \\
Representative Micro F1 & 0.78 \\
Human--Human Cohen's $\kappa$ & 0.76 \\
\hline
\end{tabular}
\caption{Gold annotation validation on a manually annotated subset.}
\label{tab:gold_validation}
\end{table}

The results demonstrate relatively strong agreement between CLARA annotations and human developmental judgments. The human--human agreement score further suggests that developmental interpretation remains partially subjective even among expert evaluators.

\section{Publisher--Educator Agreement Validation}
\label{app:publisher_validation}

Because publisher-provided developmental references constitute the primary benchmark target, we evaluate whether these references align with independent educator judgments.

We randomly sample 120 stories from the benchmark and recruit five educators with early-childhood education experience to assign developmental age groups without observing the original publisher recommendations.

\begin{table}[t]
\centering
\small
\setlength{\tabcolsep}{5pt}
\begin{tabular}{lc}
\hline
Metric & Score \\
\hline
Exact Agreement & 0.68 \\
Adjacent Agreement & 0.89 \\
Cohen's $\kappa$ & 0.72 \\
\hline
\end{tabular}
\caption{Agreement between publisher developmental references and educator judgments.}
\label{tab:publisher_validation}
\end{table}

The results show that publisher recommendations exhibit reasonably strong alignment with independent educator judgments despite ambiguity in stories involving broad developmental ranges or abstract reasoning. These findings support the use of publisher age recommendations as noisy but meaningful silver developmental references.

\section{Representative Dataset Example}
\label{app:dataset_example}

Table~\ref{tab:dataset_example_appendix} presents a representative example from the CLARA dataset.

\begin{table}[t]
\centering
\small
\renewcommand{\arraystretch}{1.15}
\setlength{\tabcolsep}{4pt}

\begin{tabular}{p{7.2cm}}
\hline

\textbf{Story Excerpt} \\
A cloth bear receives a letter inviting him to visit the big bears when the moon is full. On the journey, the bear takes a bus, climbs a mountain path, and finally joins the celebration with the other bears.
\\
\hline

\textbf{Developmental Labels} \\
\textbf{COG:} simple cause; action consequence; routine understanding

\textbf{LAN:} basic vocabulary; story sequence

\textbf{SEL:} emotion expression; preference expression; imitation
\\
\hline

\textbf{Publisher Reference} \\
2--5 (mapped to 2--4 and 4--6)
\\
\hline

\textbf{CLARA Prediction} \\
4--6
\\

\hline
\end{tabular}

\caption{Representative example from the CLARA dataset.}
\label{tab:dataset_example_appendix}

\end{table}

\section{Representative Disagreement Example}
\label{app:disagreement_case}

Table~\ref{tab:disagreement_case_appendix} presents a representative disagreement case between publisher-provided age recommendations and CLARA predictions.

\begin{table}[t]
\centering
\small
\renewcommand{\arraystretch}{1.15}
\setlength{\tabcolsep}{4pt}

\begin{tabular}{p{7.2cm}}
\hline

\textbf{Story Excerpt} \\
A man is rescued by the Nine-Colored Deer after falling into a river. Although he promises to keep the deer secret, he later betrays the deer in exchange for a reward. The story follows the consequences of betrayal, trust, and moral responsibility.
\\
\hline

\textbf{Developmental Labels} \\
\textbf{COG:} cause effect; prediction; sequence order

\textbf{LAN:} narrative inference; discourse understanding

\textbf{SEL:} emotion cause; perspective taking; rule awareness
\\
\hline

\textbf{Publisher Reference} \\
3--8 (mapped to 2--4, 4--6, 6--8)
\\
\hline

\textbf{CLARA Prediction} \\
6--8
\\

\hline
\end{tabular}

\caption{Representative disagreement case between publisher-provided age recommendations and CLARA predictions.}
\label{tab:disagreement_case_appendix}

\end{table}

\section{Human Evaluation Questionnaire}
\label{app:human_eval_questions}

Human evaluators were asked to assess stories and corresponding CLARA annotations using a 1--5 Likert scale.

\begin{itemize}

\item \textbf{COG Label Quality:}
Do the cognitive labels accurately reflect the reasoning abilities required by the story?

\item \textbf{LAN Label Quality:}
Do the language-related labels appropriately describe the language complexity of the story?

\item \textbf{SEL Label Quality:}
Do the social-emotional labels reflect the emotional and interpersonal understanding required by the story?

\item \textbf{Educational Interpretability:}
Are the developmental labels educationally meaningful and interpretable?

\item \textbf{Predicted Age Appropriateness:}
Does the predicted developmental age group reasonably match the story content?

\end{itemize}

Scores range from 1 (strongly disagree) to 5 (strongly agree).

\section{Prompt Templates}
\label{app:prompts}

\subsection{CLARA Annotation Prompt}

\begin{quote}
You are an expert in developmental psychology and children's narrative understanding.

Given a children's story, analyze the developmental abilities required for comprehension.

Assign developmental labels from the predefined taxonomy across the following dimensions:

\begin{itemize}
    \item Cognitive Skills (COG)
    \item Language Abilities (LAN)
    \item Social-Emotional Skills (SEL)
\end{itemize}

Select only labels that are strongly supported by the story content.
Avoid unsupported, redundant, or weakly related labels.

% After assigning labels, predict the most appropriate developmental age group from:
% 0--2, 2--4, 4--6, 6--8, 8--12.

% Return the results in structured JSON format.
Return the selected developmental labels in structured JSON format.
\end{quote}

\subsection{Direct Prompting Baseline}

\begin{quote}
Given a children's story, predict the most appropriate developmental age group from the following categories:

0--2, 2--4, 4--6, 6--8, 8--12.

Base your prediction on the developmental abilities required to understand the story, including reasoning complexity, language understanding, emotional interpretation, and social understanding.

Return only the predicted age group.
\end{quote}

\subsection{Rubric-Guided Prompting Baseline}

\begin{quote}
Given a children's story, first analyze the following aspects:

\begin{itemize}
    \item Cognitive complexity
    \item Language complexity
    \item Social-emotional understanding
\end{itemize}

Based on these factors, predict the most appropriate developmental age group from:

0--2, 2--4, 4--6, 6--8, 8--12.

Do not use predefined developmental labels.

Return only the predicted age group and a brief explanation.
\end{quote}

\subsection{LLM-as-Judge Prompt}

\begin{quote}
You are asked to compare two developmental age assessments for the same children's story.

One assessment is provided by the publisher.
The other assessment is generated by a structured developmental annotation system.

Evaluate which assessment better matches the developmental abilities required for understanding the story.

Consider:
\begin{itemize}
    \item reasoning complexity
    \item language understanding
    \item emotional understanding
    \item social and moral interpretation
\end{itemize}

Choose one of the following options:

\begin{itemize}
    \item Prefer CLARA
    \item Prefer Publisher
    \item Both Reasonable
\end{itemize}

Provide a brief explanation for the decision.
\end{quote}

\subsection{Human Ranking Evaluation Prompt}

\begin{quote}
Given two children's stories, determine which story requires more advanced developmental abilities for comprehension.

Consider:
\begin{itemize}
    \item reasoning complexity
    \item language understanding
    \item emotional interpretation
    \item social understanding
\end{itemize}

Choose one:
\begin{itemize}
    \item Story A requires more advanced developmental abilities
    \item Story B requires more advanced developmental abilities
    \item Both are similar
\end{itemize}
\end{quote}

\section{Full developmental label taxonomy}
\label{app:taxonomy}

Some labels intentionally preserve compact annotation-oriented naming conventions to maintain consistency across visualization figures, prompts, and annotation outputs. 
% Table~\ref{tab:taxonomy_full} presents the complete cognitive label taxonomy used in CLARA. 
Table~\ref{tab:taxonomy_full} presents the complete developmental label inventory used in CLARA.
The taxonomy contains over one hundred labels across five developmental age groups and three dimensions: COG, LAN, and SEL.

\begin{table}[t]
\centering
\scriptsize
\renewcommand{\arraystretch}{1.1}

\begin{tabular}{p{7.2cm}}
\toprule

\textbf{0--2 years (Sensorimotor stage)} \\

\midrule

\textbf{COG:}
visual tracking, sound response, face recognition, object recognition, object permanence, color recognition, shape recognition, size discrimination, pattern recognition, sensory integration
\\

\textbf{LAN:}
sound association, babbling pattern, word recognition, word-object mapping, phonetic awareness
\\

\textbf{SEL:}
emotion reaction, emotion recognition, attachment security, social smiling, caregiver response
\\

\midrule

\textbf{2--4 years (Early preoperational stage)} \\

\midrule

\textbf{COG:}
simple cause, action consequence, routine understanding, object function, basic counting, category learning, similarity detection, difference detection
\\

\textbf{LAN:}
basic vocabulary, two-word sentence, sentence imitation, naming objects, story sequence, picture description, symbolic play language
\\

\textbf{SEL:}
emotion expression, emotion labeling basic, imitation, parallel play, role play, self-awareness basic, preference expression, turn-taking basic
\\

\midrule

\textbf{4--6 years (Late preoperational stage)} \\

\midrule

\textbf{COG:}
cause effect, prediction, sequence order, temporal understanding, spatial relation, problem solution basic, categorization, attribute sorting, logical relation basic, multi attribute comparison
\\

\textbf{LAN:}
sentence comprehension, grammar structure, vocabulary expansion, simple narrative, character tracking, event linking, story retelling, dialogue understanding, basic inference, question answering
\\

\textbf{SEL:}
emotion labeling, emotion cause, emotion regulation basic, friendship, sharing behavior, cooperation, rule awareness, rule following, social norms, group participation
\\

\midrule

\textbf{6--8 years (Concrete operational stage)} \\

\midrule

\textbf{COG:}
logical reasoning,
multi-step reasoning,
causal chain understanding,
problem solving,
problem decomposition,
conservation understanding,
reversibility understanding,
hierarchical categorization
\\

\textbf{LAN:}
reading comprehension, early reading, independent reading, text inference, summarization, story structure, narrative coherence, multi event sequence
\\

\textbf{SEL:}
emotion regulation, empathy basic, perspective taking basic, conflict resolution, responsibility, cooperation advanced, self control, decision making basic
\\

\midrule

\textbf{8--12 years (Formal operational stage)} \\

\midrule

\textbf{COG:}
abstract thinking, analogy reasoning, hypothesis generation, counterfactual thinking, concept generalization, symbolic reasoning, rule abstraction, meta cognition
\\

\textbf{LAN:}
critical reading, theme understanding, figurative language, metaphor understanding, argumentation, perspective language, cross text analysis
\\

\textbf{SEL:}
empathy, perspective taking, moral reasoning, ethical reasoning, fairness, honesty, self reflection, goal setting, identity awareness, prosocial behavior
\\

\bottomrule
\end{tabular}

\caption{Full developmental label taxonomy used in CLARA.}
\label{tab:taxonomy_full}

\end{table}

\renewcommand{\arraystretch}{1.0}

\end{document}